\documentclass[conference]{IEEEtran}
\usepackage[numbers,sort&compress]{natbib}
\usepackage{multirow} 
\usepackage{xcolor}   

\usepackage{microtype}
\usepackage{graphicx}
\usepackage{subfigure}
\usepackage{booktabs}
\usepackage{amsmath,amssymb,amsfonts}
\usepackage{cite}
\usepackage{url}
\usepackage{algorithmic}
\usepackage{textcomp}
\usepackage{xcolor}
\usepackage{tabularx}
\usepackage{array}
\usepackage[compatibility=false]{caption}
\usepackage{tikz}
\usetikzlibrary{shapes.geometric, arrows, positioning, calc, arrows.meta}
\usepackage{tikz-network}
\usepackage[capitalize,noabbrev]{cleveref}

\usepackage[table]{xcolor}
\usepackage{xspace}
\definecolor{headerblue}{RGB}{41,98,255}
\definecolor{sectionbg}{RGB}{230,240,255}
\definecolor{zebra}{RGB}{248,249,250}
\definecolor{avgcolor}{RGB}{255,241,118}
\definecolor{good}{RGB}{0,128,0}
\definecolor{bad}{RGB}{220,20,60}
\usepackage{fontawesome5} 
\definecolor{headerblue}{RGB}{41,98,255}
\definecolor{zebra}{RGB}{248,249,250}
\usepackage{pifont}

\def\BibTeX{{\rm B\kern-.05em{\sc i\kern-.025em b}\kern-.08em
    T\kern-.1667em\lower.7ex\hbox{E}\kern-.125emX}}
 \IEEEoverridecommandlockouts
\begin{document}

\title{Leveraging Large Language Models for Sentiment Analysis: Multi-Modal Analysis of Decentraland's MANA Token}

\author{
\IEEEauthorblockN{Xintong Wu}
\IEEEauthorblockA{
\textit{University of Pennsylvania}\\ 
Philadelphia, PA, USA 
}

\and

\IEEEauthorblockN{Peiting Tsai, Jing Yuan, Michael Yu, and Greg Sun}
\IEEEauthorblockA{
\textit{Microsoft}\\ 
Beijing, China
}

\and

\IEEEauthorblockN{Luyao Zhang\textsuperscript{*}}
\IEEEauthorblockA{
\textit{Duke Kunshan University}\\ 
Kunshan, Jiangsu, China}

\thanks{\textsuperscript{*}Corresponding author: Luyao Zhang (lz183@duke.edu), Digital Innovation Research Center and Social Science Division, Duke Kunshan University. Address: Duke Avenue No.8, Kunshan, Suzhou, Jiangsu, China, 215316. \textbf{Acknowledgments}: Luyao Zhang acknowledges support from the National Science Foundation of China (NSFC) for the project titled "Trust Mechanism Design on Blockchain: An Interdisciplinary Approach of Game Theory, Reinforcement Learning, and Human-AI Interactions" (Grant No. 12201266). Xintong Wu is grateful for the support provided by the Office of Academic Services, the Summer Research Scholar Program, and the Student Experiential Learning Fellow (SELF) Program at Duke Kunshan University, supervised by Prof. Luyao Zhang. Xintong also extends appreciation to all members of the One-Person Entrepreneur (OPE) group at Microsoft China for their valuable inspiration.}}

\maketitle

\begin{abstract}
Decentraland, a decentralized virtual reality platform operating within the expanding Metaverse ecosystem, utilizes its native MANA token to facilitate virtual asset transactions and governance. This study investigates the integration of Discord community sentiment with multi-modal financial data to enhance cryptocurrency price prediction within virtual world economies. We address: (1) identifying sentiment patterns within Decentraland's Discord community, and (2) evaluating the impact of multi-modal features on token return forecasting. Using a BERT-based large language model for sentiment analysis, we develop two LSTM architectures: a baseline incorporating historical prices and a multi-modal variant integrating sentiment scores, trading volume, and market capitalization. Results indicate predominantly neutral community sentiment with a positive skew. The multi-modal model significantly outperforms the price-only baseline in prediction accuracy. These findings demonstrate the predictive value of community-derived signals for virtual economy forecasting and establish a foundation for future research at the intersection of immersive virtual environments, natural language processing, and cryptocurrency market analysis.
\end{abstract}

\begin{IEEEkeywords}
Large Language Models, Sentiment Analysis, Multi-Modal Learning, Cryptocurrency Prediction, Metaverse, Blockchain Economics 
\end{IEEEkeywords}

\section{Introduction}

The emergence of blockchain-enabled immersive virtual environments—the Metaverse—has catalyzed interest in persistent digital economies where users transact virtual assets through native cryptocurrencies \citep{huang2024scalability,truong2023blockchain}. Decentraland represents a paradigmatic case of this convergence, combining virtual reality with decentralized infrastructure to create economic spheres with unique market dynamics \citep{goanta2020selling}. Its native token, MANA, facilitates transactions for virtual goods and land parcels while conferring governance rights through a decentralized autonomous organization (DAO), rendering it ideal for examining sentiment dynamics in nascent virtual economies.
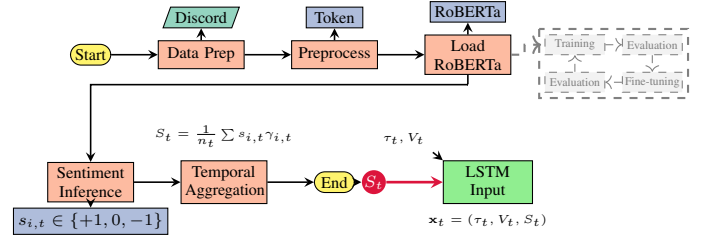
\begin{figure}[!t]
    \centering
    \begin{tikzpicture}[
        node distance=0.6cm and 0.7cm, 
        every node/.style={font=\scriptsize},
        scale=0.88, 
        transform shape
    ]
        
        \definecolor{myblue}{RGB}{141,160,203}
        \definecolor{mygreen}{RGB}{102,194,165}
        \definecolor{myorange}{RGB}{251,141,98}
        \definecolor{lstmcolor}{RGB}{144,238,144}
        \definecolor{highlight}{RGB}{220,20,60}

        \tikzstyle{startstop} = [rectangle, rounded corners, minimum width=0.6cm, minimum height=0.28cm, text centered, draw=black, fill=yellow!70, inner sep=2pt]
        \tikzstyle{io} = [trapezium, trapezium stretches=true, trapezium left angle=70, trapezium right angle=110, minimum width=0.8cm, minimum height=0.25cm, text centered, draw=black, fill=mygreen!70, inner sep=2pt]
        \tikzstyle{process} = [rectangle, minimum width=1.15cm, minimum height=0.42cm, text centered, text width=1.15cm, draw=black, fill=myorange!60, inner sep=2pt]
        \tikzstyle{process1} = [rectangle, minimum width=0.85cm, minimum height=0.25cm, text centered, draw=black, fill=myblue!70, inner sep=2pt]
        \tikzstyle{lstmnode} = [rectangle, minimum width=1.2cm, minimum height=0.42cm, text centered, text width=1.2cm, draw=black, fill=lstmcolor, inner sep=2pt]
        \tikzstyle{arrow} = [semithick,->,>=stealth]

        \tikzstyle{trainnode} = [rectangle, dashed, draw=gray, fill=gray!10, text=gray, minimum width=0.9cm, minimum height=0.30cm, text centered, inner sep=1pt, font=\tiny]

        \node (start) [startstop] {Start};
        \node (dataPrep) [process, right=of start] {Data Prep};
        \node (dataPre) [process, right=of dataPrep] {Preprocess};
        \node (modelLoad) [process, right=of dataPre] {Load RoBERTa};

        \node (sentimentInf) [process, below=of start, yshift=-0.85cm] {Sentiment\\Inference};
        \node (aggregation) [process, right=of sentimentInf] {Temporal\\Aggregation};
        \node (end) [startstop, right=of aggregation] {End};
        \node (stNode) [circle, fill=highlight, text=white, minimum size=0.36cm, inner sep=0pt, right=0.1cm of end] {$S_t$};
        \node (lstm) [lstmnode, right=of stNode, xshift=0.15cm] {LSTM Input};

        \node (dailyFormula) [font=\tiny, above=0.08cm of aggregation] {$S_t=\frac{1}{n_t}\sum s_{i,t}\gamma_{i,t}$};

        \node (modelTrain) [trainnode, right=0.5cm of modelLoad, yshift=0.15cm] {Training};
        \node (eval1) [trainnode, right=0.25cm of modelTrain] {Evaluation};
        \node (finetune) [trainnode, below=0.25cm of eval1] {Fine-tuning};
        \node (eval2) [trainnode, left=0.25cm of finetune] {Evaluation};

        \draw [dashed, gray, ->] (modelTrain) -- (eval1);
        \draw [dashed, gray, ->] (eval1) -- (finetune);
        \draw [dashed, gray, ->] (finetune) -- (eval2);
        \draw [dashed, gray, ->] (eval2) -- (modelTrain);

        \draw [dashed, gray, thick, ->] (modelLoad.east) -- ++(0.2,0) |- ($(modelTrain.west)-(0.05,0)$);

        \draw [dashed, gray, thick] 
            ($(modelTrain.north west)+(-0.08,0.08)$) 
            rectangle 
            ($(finetune.south east)+(0.08,-0.08)$);

        \node (textInput) [io, above=0.18cm of dataPrep] {Discord};
        \node (tokenization) [process1, above=0.18cm of dataPre] {Token};
        \node (roberta) [process1, above=0.18cm of modelLoad] {RoBERTa};
        
        \node (scoring) [process1, below=0.08cm of sentimentInf] {$s_{i,t}\in\{+1,0,-1\}$};

        \node (priceVol) [above left=0.12cm and 0.15cm of lstm.north west, font=\tiny] {$\tau_t, V_t$};
        \draw [arrow, dashed] (priceVol.south east) -- (lstm.north west);

        \draw [arrow] (start) -- (dataPrep);
        \draw [arrow] (dataPrep) -- (dataPre);
        \draw [arrow] (dataPre) -- (modelLoad);
        \draw [arrow] (modelLoad) |- ++(0,-0.45) -| (sentimentInf);
        \draw [arrow] (sentimentInf) -- (aggregation);
        \draw [arrow] (aggregation) -- (end);
        \draw [arrow] (end) -- (stNode);
        \draw [arrow, line width=1.0pt, color=highlight] (stNode) -- (lstm);
        
        \draw [arrow] (dataPrep) -- (textInput);
        \draw [arrow] (dataPre) -- (tokenization);
        \draw [arrow] (modelLoad) -- (roberta);
        \draw [arrow] (sentimentInf) -- (scoring);
        
        \node [below=0.03cm of lstm, font=\tiny] {$\mathbf{x}_t=(\tau_t,V_t,S_t)$};

    \end{tikzpicture}
    \vspace{-0.1cm}
    \captionsetup{font=footnotesize}
    \caption{LLM-based sentiment analysis pipeline. \textit{Red circle highlights sentiment output $S_t$ feeding into downstream LSTM alongside conventional technical indicators such as price $\tau_t$ and volume $V_t$.}}
    \label{fig:sentiment_analysis_flowchart}
\end{figure}
Investor sentiment from social media constitutes a significant predictor of market behavior, prompting extensive applications of sentiment analysis in financial forecasting \citep{guidi2022social}. Within Decentraland, Discord serves as the primary communication channel where user discussions may influence MANA price dynamics. Recent advances in large language models (LLMs) have enhanced sentiment analysis capabilities, with multi-agent frameworks improving financial classification accuracy \citep{10.1145/3688399} and specialized models advancing emotion detection \citep{10.1145/3637528.3671552}.

This study constructs a multi-modal dataset integrating historical price data with Discord-derived sentiment, trading volume, and market capitalization to forecast token returns. We investigate two research questions (RQs): 
\begin{itemize}
    \item \textbf{(RQ1)} What are the prevailing sentiment patterns among Decentraland Discord users, and to what extent do these reflect platform dynamics? 
    \item \textbf{(RQ2)} How does integrating multi-modal features affect token return prediction accuracy relative to price-only models?
\end{itemize}

Employing a BERT-based architecture for sentiment extraction and LSTM networks for time-series prediction, we evaluate the incremental value of community sentiment for cryptocurrency forecasting. Figure~\ref{fig:sentiment_analysis_flowchart} illustrates the LLM-based sentiment analysis pipeline.\footnote{All data and code are publicly accessible at: \url{https://github.com/Xintong1122/Decentraland_Senti}}
\section{Data}

\subsection{Data Collection}

\subsubsection{Numerical Data}
Daily OHLCV data for MANA (ERC-20 utility token \citep{decentraland2024whitepaper, decentraland2017whitepaper, hu2025exploring}) spanning $t \in [1,T]$ where $T=366$ days (June 15, 2023--June 14, 2024) comprise open, high, low, close prices ($P_t^O, P_t^H, P_t^L, P_t^C$), trading volume $V_t$, and market capitalization $M_t$ (USD), obtained from \textit{CoinMarketCap}.\footnote{\url{https://coinmarketcap.com/}}

\subsubsection{Textual Data} 
Discord \textit{\#general} channel messages $\mathcal{M}=\{m_i\}_{i=1}^n$ ($n=5,883$) were harvested via \textit{DiscordChatExporter},\footnote{\url{https://github.com/Tyrrrz/DiscordChatExporter}} yielding timestamps $t_i$, anonymized author IDs $\alpha_i$, content $c_i$, and engagement metrics. Post-filtering invalid entries, $n'=5,513$ messages ($\mathcal{M}' \subset \mathcal{M}$) comprised the final corpus.

\subsection{Data Processing}

\subsubsection{Price Metric}
The typical price $\tau_t = \frac{1}{3}(P_t^H + P_t^L + P_t^C)$ serves as the daily indicator, smoothing intraday fluctuations following standard technical analysis practice \citep{murphy1999technical}.

\subsubsection{Return Calculation}
Continuous returns $r_t  = \ln(\tau_{t+1}/\tau_t) \\= \ln(\tau_{t+1}) - \ln(\tau_t)$ stabilize variance and ensure temporal comparability for ERC-20 token markets \citep{heinonen2020collective}.

\section{Methodology}
\subsection{Sentiment Analysis}
\subsubsection{LLM Selection}
We employ the RoBERTa-based classifier \textit{cardiffnlp/twitter-roberta-base-sentiment-latest}\footnote{https://huggingface.co/cardiffnlp/twitter-roberta-base-sentiment-latest} from Hugging Face, pretrained on social media corpora for informal text classification \citep{devlin2018bert, masala2020robert}.

\subsubsection{Aggregation}
Each message $i$ at time $t$ maps to sentiment $s_{i,t} \in \{+1, 0, -1\}$ (positive, neutral, negative) with confidence $\gamma_{i,t}$. Daily sentiment aggregates as $S_t = \frac{1}{n_t}\sum_{i=1}^{n_t} s_{i,t}\gamma_{i,t}$, where $n_t$ denotes daily message volume \citep{zhang2023enhancing}. Discretization thresholds define $S_t \in [-1,-0.1)$ as negative, $[-0.1,0.1]$ as neutral, and $(0.1,1]$ as positive.

\subsection{Return Prediction}
\subsubsection{LSTM Architecture}
Long Short-Term Memory networks model sequential dependencies $\{(\mathbf{x}_t, r_t)\}_{t=1}^T$ via gated mechanisms \citep{yu2019review,sherstinsky2020fundamentals,staudemeyer2019understanding}:
\begin{equation}
i_t = \sigma(W_i \cdot [h_{t-1}, \mathbf{x}_t] + b_i), \quad
o_t = \sigma(W_o \cdot [h_{t-1}, \mathbf{x}_t] + b_o)
\end{equation}
where $i_t, o_t$ denote input and output gates, $\sigma$ the sigmoid function, and $W, b$ learned parameters. Model specifications:
\begin{itemize}
\item \textbf{Baseline:} $\mathbf{x}_t = \tau_t$ (scalar, typical price only)
\item \textbf{Multi-modal:} $\mathbf{x}_t = (\tau_t, V_t, M_t, S_t)' \in \mathcal{R}^4$ 
\end{itemize}

\subsubsection{Evaluation Metrics}
Predictive performance is quantified via $\text{MSE} = T^{-1}\sum_{t=1}^T (r_t - \hat{r}_t)^2$, $\text{MAE} = T^{-1}\sum_{t=1}^T |r_t - \hat{r}_t|$, and $R^2 = 1 - {\sum_{t=1}^T (r_t - \hat{r}_t)^2}/{\sum_{t=1}^T (r_t - \bar{r})^2}$ \citep{chicco2021coefficient,koksoy2006multiresponse,willmott2005advantages,chai2014root,cameron1997r}.

\section{Results}

\begin{figure}[htbp]
\centering
\includegraphics[width=1\linewidth]{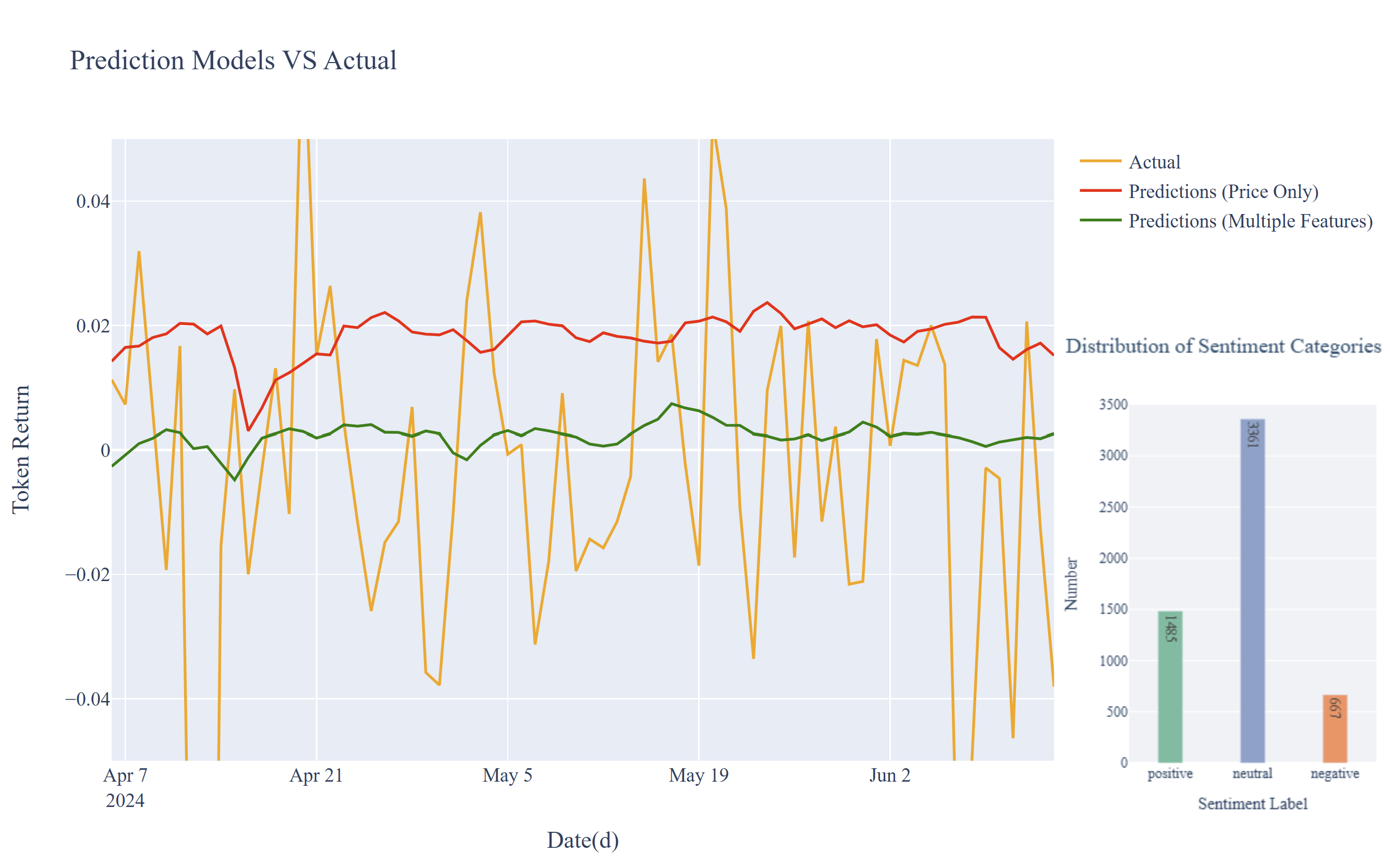}
\captionsetup{skip=0.1cm}
\captionsetup{font=footnotesize}
\caption{Price Prediction and Sentiment Categories.}
\vspace{-0.5cm}
\label{fig:new}
\end{figure}

Figure~\ref{fig:new} presents the dual analysis of community sentiment dynamics and predictive model performance. The bar plot illustrates the distribution of Discord sentiment over the 366-day observation period: 26.94\% positive ($s_{i,t}=+1$), 60.96\% neutral ($s_{i,t}=0$), and 12.10\% negative ($s_{i,t}=-1$). This predominance of neutral sentiment with a positive skew indicates stable community affect, with daily aggregate scores $S_t$ concentrating within $[0,1]$. Notably, negative sentiment spikes correlate primarily with user-level technical issues rather than exogenous market events \citep{decentralandfoundation_2023,gibbons_2024}, suggesting these data reflect platform user experience rather than macroeconomic signals.

The time-series plots compare the baseline LSTM (price-only feature $\mathbf{x}_t = \tau_t$) against the multi-modal architecture ($\mathbf{x}_t = (\tau_t, V_t, M_t, S_t)' \in \mathcal{R}^4$). The multi-modal model improves prediction accuracy, reducing $\text{MSE}$ from $0.002100$ to $0.001528$ and $\text{MAE}$ from $0.0297$ to $0.0241$. However, both models exhibit negative $R^2$ values (baseline: $-0.4185$; multi-modal: $-0.0321$), indicating limited explanatory power for directional trends. While auxiliary features enhance short-term precision, they inadequately capture broader market dynamics, consistent with \citep{lee2024test}.

\section{Conclusion}

Our study contributes to metaverse and blockchain-empowered virtual economy research \citep{ duan2021metaverse, guidi2022social, le2024blockchain, wu2022metaverse, den2023metaverse, zhang2024nft, yan2025ethereum, chemaya2025uniswap} and the literature on user-generated content and community sentiment \citep{cheong2008consumers, bahtar2016impact, micu2017analyzing, yang2015twitter}. By applying large language models to financial sentiment \citep{kasneci2023chatgpt, chang2024survey, huang2023chatgpt, li2018deep, liu2019roberta, xu2019bert, zhan2024optimization, tornberg2023simulating, deng2023llms}, we extend methodological approaches in sentiment analysis \citep{medhat2014sentiment, wankhade2022survey, hussein2018survey, devika2016sentiment, gonccalves2013comparing, yue2019survey, neri2012sentiment} to the cryptocurrency domain \citep{elbahrawy2017evolutionary, henrique2019literature, chen2019sentiment, zimbra2018state, melton2021public, alharbi2019twitter, quan2023decoding}. Results show that incorporating sentiment enhances price-only models \citep{siami2018forecasting, athey2019machine, hamilton2020time}, though gains remain limited as Discord sentiment reflects platform-specific user experience rather than macro-market conditions. Future work should integrate additional data sources \citep{li2020multimodal}, improve domain-specific modeling, and consider scalability and security aspects \citep{huang2024scalability}.\footnote{Detailed contributions to each literature stream are provided in Appendix~\ref{app:contributions}.}

\bibliographystyle{IEEEtran}
\bibliography{reference}

@techreport{decentraland2024whitepaper,
  title={Decentraland White Paper 2.0},
  author={{Decentraland Foundation}},
  year={2024},
  url={https://decentraland.org/whitepaper2.pdf},
  institution={Decentraland Foundation},
  note={Accessed: 2024}
}

@techreport{decentraland2017whitepaper,
  title={Decentraland: A Blockchain-Based Virtual World},
  author={Ordano, Esteban and Meilich, Ariel and Jardi, Yemel and Araoz, Manuel},
  year={2017},
  url={https://decentraland.org/whitepaper.pdf},
  institution={Decentraland}
}

@article{hu2025exploring,
  title={Exploring the Decentraland Economy: Multifaceted Parcel Attributes, Key Insights, and Benchmarking},
  author={Hu, Yuntao and Li, Zhenyu and Chen, Jiaqi and Gao, Yue and Liu, Yong},
  journal={arXiv preprint arXiv:2404.07533},
  year={2025},
  url={https://arxiv.org/abs/2404.07533}
}

@inproceedings{zhang2024nft,
  author={Zhang, Luyao and Quan, Yutong and Cao, Jiaxun and Zhou, Kyrie Zhixuan and Tong, Xin},
  booktitle={2024 IEEE 24th International Conference on Software Quality, Reliability, and Security Companion (QRS-C)}, 
  title={Leveraging Social Media Sentiments and Ethical Signals for {NFT} Valuation}, 
  year={2024},
  pages={206--215},
  publisher={IEEE},
  doi={10.1109/QRS-C63300.2024.00036},
  url={https://ieeexplore.ieee.org/document/10727085}
}

@article{yan2025ethereum,
  title={A Data Engineering Framework for Ethereum Beacon Chain Rewards: From Data Collection to Decentralization Metrics},
  author={Yan, Tao and Li, Shengman and Kramer, Benjamin and Zhang, Luyao and Tessone, Claudio J},
  journal={Scientific Data},
  volume={12},
  number={1},
  pages={519},
  year={2025},
  publisher={Springer Nature},
  doi={10.1038/s41597-025-04623-7},
  url={https://www.nature.com/articles/s41597-025-04623-7}
}

@article{chemaya2025uniswap,
  title={A Dataset of Uniswap Daily Transaction Indices by Network},
  author={Chemaya, Nir and Cong, Lin William and Joergensen, Emma and Liu, Dingyue and Zhang, Luyao},
  journal={Scientific Data},
  volume={12},
  number={1},
  pages={93},
  year={2025},
  publisher={Springer Nature},
  doi={10.1038/s41597-024-04042-0},
  url={https://www.nature.com/articles/s41597-024-04042-0}
}

@book{murphy1999technical,
  title={Technical analysis of the financial markets: A comprehensive guide to trading methods and applications},
  author={Murphy, John J},
  year={1999},
  publisher={Penguin}
}

@inproceedings{guidi2022social,
  title={Social games and Blockchain: exploring the Metaverse of Decentraland},
  author={Guidi, Barbara and Michienzi, Andrea},
  booktitle={2022 IEEE 42nd International Conference on Distributed Computing Systems Workshops (ICDCSW)},
  pages={199--204},
  year={2022},
  organization={IEEE}
}

@inproceedings{heinonen2020collective,
  title={Collective behavior of price changes of erc-20 tokens},
  author={Heinonen, Henri T and Semenov, Alexander and Boginski, Vladimir},
  booktitle={Computational Data and Social Networks: 9th International Conference, CSoNet 2020, Dallas, TX, USA, December 11--13, 2020, Proceedings 9},
  pages={487--498},
  year={2020},
  organization={Springer}
}

@inproceedings{devlin2018bert,
  title={BERT: Pre-training of Deep Bidirectional Transformers for Language Understanding},
  author={Devlin, Jacob and Chang, Ming-Wei and Lee, Kenton and Toutanova, Kristina},
  booktitle={Proceedings of the 2019 Conference of the North American Chapter of the Association for Computational Linguistics: Human Language Technologies, Volume 1 (Long and Short Papers)},
  pages={4171--4186},
  year={2019},
  publisher={Association for Computational Linguistics},
  url={https://aclanthology.org/N19-1423/},
  doi={10.18653/v1/N19-1423}
}

@inproceedings{masala2020robert,
  title={Robert--a romanian bert model},
  author={Masala, Mihai and Ruseti, Stefan and Dascalu, Mihai},
  booktitle={Proceedings of the 28th International Conference on Computational Linguistics},
  pages={6626--6637},
  year={2020}
}

@inproceedings{zhang2023enhancing,
  title={Enhancing financial sentiment analysis via retrieval augmented large language models},
  author={Zhang, Boyu and Yang, Hongyang and Zhou, Tianyu and Ali Babar, Muhammad and Liu, Xiao-Yang},
  booktitle={Proceedings of the fourth ACM international conference on AI in finance},
  pages={349--356},
  year={2023}
}

@article{yu2019review,
  title={A review of recurrent neural networks: LSTM cells and network architectures},
  author={Yu, Yong and Si, Xiaosheng and Hu, Changhua and Zhang, Jianxun},
  journal={Neural computation},
  volume={31},
  number={7},
  pages={1235--1270},
  year={2019},
  publisher={MIT Press One Rogers Street, Cambridge, MA 02142-1209, USA journals-info~…}
}

@article{staudemeyer2019understanding,
  title={Understanding LSTM--a tutorial into long short-term memory recurrent neural networks},
  author={Staudemeyer, Ralf C and Morris, Eric Rothstein},
  journal={arXiv preprint arXiv:1909.09586},
  year={2019}
}

@article{sherstinsky2020fundamentals,
  title={Fundamentals of recurrent neural network (RNN) and long short-term memory (LSTM) network},
  author={Sherstinsky, Alex},
  journal={Physica D: Nonlinear Phenomena},
  volume={404},
  pages={132306},
  year={2020},
  publisher={Elsevier}
}

@article{chicco2021coefficient,
  title={The coefficient of determination R-squared is more informative than SMAPE, MAE, MAPE, MSE and RMSE in regression analysis evaluation},
  author={Chicco, Davide and Warrens, Matthijs J and Jurman, Giuseppe},
  journal={Peerj computer science},
  volume={7},
  pages={e623},
  year={2021},
  publisher={PeerJ Inc.}
}

@article{koksoy2006multiresponse,
  title={Multiresponse robust design: Mean square error (MSE) criterion},
  author={K{\"o}ksoy, Onur},
  journal={Applied Mathematics and Computation},
  volume={175},
  number={2},
  pages={1716--1729},
  year={2006},
  publisher={Elsevier}
}

@article{willmott2005advantages,
  title={Advantages of the mean absolute error (MAE) over the root mean square error (RMSE) in assessing average model performance},
  author={Willmott, Cort J and Matsuura, Kenji},
  journal={Climate research},
  volume={30},
  number={1},
  pages={79--82},
  year={2005}
}

@article{chai2014root,
  title={Root mean square error (RMSE) or mean absolute error (MAE)?--Arguments against avoiding RMSE in the literature},
  author={Chai, Tianfeng and Draxler, Roland R},
  journal={Geoscientific model development},
  volume={7},
  number={3},
  pages={1247--1250},
  year={2014},
  publisher={Copernicus Publications G{\"o}ttingen, Germany}
}

@article{cameron1997r,
  title={An R-squared measure of goodness of fit for some common nonlinear regression models},
  author={Cameron, A Colin and Windmeijer, Frank AG},
  journal={Journal of econometrics},
  volume={77},
  number={2},
  pages={329--342},
  year={1997},
  publisher={Elsevier}
}

@misc{gibbons_2024, title={Decentraland Art Week 2024 Kicks Off}, url={https://www.blockleaders.io/news/decentraland-art-week-2024-kicks-off}, journal={Blockchain Leaders}, author={Gibbons, Lisa }, year={2024}, month={Mar} }

@misc{decentralandfoundation_2023, title={Decentraland’s Weekly Newsletter - Jan 4}, url={https://decentraland.beehiiv.com/p/weekly-newsletter-january-4}, journal={Decentraland Weekly}, author={Decentraland Foundation}, year={2023} }

@article{goanta2020selling,
  title={Selling LAND in Decentraland: The regime of non-fungible tokens on the Ethereum blockchain under the digital content directive},
  author={Goanta, Catalina},
  journal={Disruptive technology, legal innovation, and the future of real estate},
  pages={139--154},
  year={2020},
  publisher={Springer}
}

@incollection{den2023metaverse,
  title={Metaverse in investment using sentiment analysis and machine learning},
  author={Den Yeoh, Eik and Chung, Tinfah and Wang, Yuyang},
  booktitle={Strategies and Opportunities for Technology in the Metaverse World},
  pages={78--113},
  year={2023},
  publisher={IGI Global}
}

@article{medhat2014sentiment,
  title={Sentiment analysis algorithms and applications: A survey},
  author={Medhat, Walaa and Hassan, Ahmed and Korashy, Hoda},
  journal={Ain Shams engineering journal},
  volume={5},
  number={4},
  pages={1093--1113},
  year={2014},
  publisher={Elsevier}
}

@article{liu2019roberta,
  title={Roberta: A robustly optimized bert pretraining approach},
  author={Liu, Yinhan and Ott, Myle and Goyal, Naman and Du, Jingfei and Joshi, Mandar and Chen, Danqi and Levy, Omer and Lewis, Mike and Zettlemoyer, Luke and Stoyanov, Veselin},
  journal={arXiv preprint arXiv:1907.11692},
  year={2019}
}

@article{tornberg2023simulating,
  title={Simulating social media using large language models to evaluate alternative news feed algorithms},
  author={T{\"o}rnberg, Petter and Valeeva, Diliara and Uitermark, Justus and Bail, Christopher},
  journal={arXiv preprint arXiv:2310.05984},
  year={2023}
}

@inproceedings{deng2023llms,
  title={What do llms know about financial markets? a case study on reddit market sentiment analysis},
  author={Deng, Xiang and Bashlovkina, Vasilisa and Han, Feng and Baumgartner, Simon and Bendersky, Michael},
  booktitle={Companion Proceedings of the ACM Web Conference 2023},
  pages={107--110},
  year={2023}
}

@article{chang2024survey,
  title={A survey on evaluation of large language models},
  author={Chang, Yupeng and Wang, Xu and Wang, Jindong and Wu, Yuan and Yang, Linyi and Zhu, Kaijie and Chen, Hao and Yi, Xiaoyuan and Wang, Cunxiang and Wang, Yidong and others},
  journal={ACM Transactions on Intelligent Systems and Technology},
  volume={15},
  number={3},
  pages={1--45},
  year={2024},
  publisher={ACM New York, NY}
}

@article{kasneci2023chatgpt,
  title={ChatGPT for good? On opportunities and challenges of large language models for education},
  author={Kasneci, Enkelejda and Se{\ss}ler, Kathrin and K{\"u}chemann, Stefan and Bannert, Maria and Dementieva, Daryna and Fischer, Frank and Gasser, Urs and Groh, Georg and G{\"u}nnemann, Stephan and H{\"u}llermeier, Eyke and others},
  journal={Learning and individual differences},
  volume={103},
  pages={102274},
  year={2023},
  publisher={Elsevier}
}

@article{huang2023chatgpt,
  title={ChatGPT for shaping the future of dentistry: the potential of multi-modal large language model},
  author={Huang, Hanyao and Zheng, Ou and Wang, Dongdong and Yin, Jiayi and Wang, Zijin and Ding, Shengxuan and Yin, Heng and Xu, Chuan and Yang, Renjie and Zheng, Qian and others},
  journal={International Journal of Oral Science},
  volume={15},
  number={1},
  pages={29},
  year={2023},
  publisher={Nature Publishing Group UK London}
}

@article{li2018deep,
  title={Deep learning for natural language processing: advantages and challenges},
  author={Li, Hang},
  journal={National Science Review},
  volume={5},
  number={1},
  pages={24--26},
  year={2018},
  publisher={Oxford University Press}
}

@inproceedings{duan2021metaverse,
  title={Metaverse for social good: A university campus prototype},
  author={Duan, Haihan and Li, Jiaye and Fan, Sizheng and Lin, Zhonghao and Wu, Xiao and Cai, Wei},
  booktitle={Proceedings of the 29th ACM international conference on multimedia},
  pages={153--161},
  year={2021}
}

@incollection{wu2022metaverse,
  title={Metaverse: The world reimagined},
  author={Wu, Brian and Wu, Bridget},
  booktitle={Blockchain for teens: With case studies and examples of blockchain across various industries},
  pages={267--313},
  year={2022},
  publisher={Springer}
}

@article{cheong2008consumers,
  title={Consumers’ reliance on product information and recommendations found in UGC},
  author={Cheong, Hyuk Jun and Morrison, Margaret A},
  journal={Journal of interactive advertising},
  volume={8},
  number={2},
  pages={38--49},
  year={2008},
  publisher={Taylor \& Francis}
}

@article{bahtar2016impact,
  title={The impact of User--Generated Content (UGC) on product reviews towards online purchasing--A conceptual framework},
  author={Bahtar, Azlin Zanariah and Muda, Mazzini},
  journal={Procedia Economics and Finance},
  volume={37},
  pages={337--342},
  year={2016},
  publisher={Elsevier}
}

@article{micu2017analyzing,
  title={Analyzing user sentiment in social media: Implications for online marketing strategy},
  author={Micu, Adrian and Micu, Angela Eliza and Geru, Marius and Lixandroiu, Radu Constantin},
  journal={Psychology \& Marketing},
  volume={34},
  number={12},
  pages={1094--1100},
  year={2017},
  publisher={Wiley Online Library}
}

@article{yang2015twitter,
  title={Twitter financial community sentiment and its predictive relationship to stock market movement},
  author={Yang, Steve Y and Mo, Sheung Yin Kevin and Liu, Anqi},
  journal={Quantitative Finance},
  volume={15},
  number={10},
  pages={1637--1656},
  year={2015},
  publisher={Taylor \& Francis}
}

@article{wankhade2022survey,
  title={A survey on sentiment analysis methods, applications, and challenges},
  author={Wankhade, Mayur and Rao, Annavarapu Chandra Sekhara and Kulkarni, Chaitanya},
  journal={Artificial Intelligence Review},
  volume={55},
  number={7},
  pages={5731--5780},
  year={2022},
  publisher={Springer}
}

@article{hussein2018survey,
  title={A survey on sentiment analysis challenges},
  author={Hussein, Doaa Mohey El-Din Mohamed},
  journal={Journal of King Saud University-Engineering Sciences},
  volume={30},
  number={4},
  pages={330--338},
  year={2018},
  publisher={Elsevier}
}

@article{devika2016sentiment,
  title={Sentiment analysis: a comparative study on different approaches},
  author={Devika, M D{\textordfeminine} and Sunitha, C{\textordfeminine} and Ganesh, Amal},
  journal={Procedia Computer Science},
  volume={87},
  pages={44--49},
  year={2016},
  publisher={Elsevier}
}

@inproceedings{gonccalves2013comparing,
  title={Comparing and combining sentiment analysis methods},
  author={Gon{\c{c}}alves, Pollyanna and Ara{\'u}jo, Matheus and Benevenuto, Fabr{\'\i}cio and Cha, Meeyoung},
  booktitle={Proceedings of the first ACM conference on Online social networks},
  pages={27--38},
  year={2013}
}

@inproceedings{xu2019bert,
  title={BERT Post-Training for Review Reading Comprehension and Aspect-Based Sentiment Analysis},
  author={Xu, Hu and Liu, Bing and Shu, Lei and Yu, Philip S},
  booktitle={Proceedings of the 2019 Conference of the North American Chapter of the Association for Computational Linguistics: Human Language Technologies, Volume 1 (Long and Short Papers)},
  pages={2324--2335},
  year={2019},
  publisher={Association for Computational Linguistics},
  url={https://aclanthology.org/N19-1242/},
  doi={10.18653/v1/N19-1242}
}

@article{zhan2024optimization,
  title={Optimization Techniques for Sentiment Analysis Based on LLM (GPT-3)},
  author={Zhan, Tong and Shi, Chenxi and Shi, Yadong and Li, Huixiang and Lin, Yiyu},
  journal={arXiv preprint arXiv:2405.09770},
  year={2024}
}

@article{yue2019survey,
  title={A survey of sentiment analysis in social media},
  author={Yue, Lin and Chen, Weitong and Li, Xue and Zuo, Wanli and Yin, Minghao},
  journal={Knowledge and Information Systems},
  volume={60},
  pages={617--663},
  year={2019},
  publisher={Springer}
}

@inproceedings{neri2012sentiment,
  title={Sentiment analysis on social media},
  author={Neri, Federico and Aliprandi, Carlo and Capeci, Federico and Cuadros, Montserrat},
  booktitle={2012 IEEE/ACM international conference on advances in social networks analysis and mining},
  pages={919--926},
  year={2012},
  organization={IEEE}
}

@article{zimbra2018state,
  title={The state-of-the-art in Twitter sentiment analysis: A review and benchmark evaluation},
  author={Zimbra, David and Abbasi, Ahmed and Zeng, Daniel and Chen, Hsinchun},
  journal={ACM Transactions on Management Information Systems (TMIS)},
  volume={9},
  number={2},
  pages={1--29},
  year={2018},
  publisher={ACM New York, NY, USA}
}

@article{melton2021public,
  title={Public sentiment analysis and topic modeling regarding COVID-19 vaccines on the Reddit social media platform: A call to action for strengthening vaccine confidence},
  author={Melton, Chad A and Olusanya, Olufunto A and Ammar, Nariman and Shaban-Nejad, Arash},
  journal={Journal of Infection and Public Health},
  volume={14},
  number={10},
  pages={1505--1512},
  year={2021},
  publisher={Elsevier}
}

@article{alharbi2019twitter,
  title={Twitter sentiment analysis with a deep neural network: An enhanced approach using user behavioral information},
  author={Alharbi, Ahmed Sulaiman M and de Doncker, Elise},
  journal={Cognitive Systems Research},
  volume={54},
  pages={50--61},
  year={2019},
  publisher={Elsevier}
}

@article{chen2019sentiment,
  title={Sentiment-induced bubbles in the cryptocurrency market},
  author={Chen, Cathy Yi-Hsuan and Hafner, Christian M},
  journal={Journal of Risk and Financial Management},
  volume={12},
  number={2},
  pages={53},
  year={2019},
  publisher={MDPI}
}

@inproceedings{quan2023decoding,
  title={Decoding Social Sentiment in DAO: A Comparative Analysis of Blockchain Governance Communities},
  author={Quan, Yutong and Wu, Xintong and Deng, Wanlin and Zhang, Luyao},
  booktitle={2024 IEEE 24th International Conference on Software Quality, Reliability, and Security Companion (QRS-C)},
  pages={216--224},
  year={2024},
  publisher={IEEE},
  doi={10.1109/QRS-C63300.2024.00037},
  url={https://doi.org/10.1109/QRS-C63300.2024.00037}
}

@article{elbahrawy2017evolutionary,
  title={Evolutionary dynamics of the cryptocurrency market},
  author={ElBahrawy, Abeer and Alessandretti, Laura and Kandler, Anne and Pastor-Satorras, Romualdo and Baronchelli, Andrea},
  journal={Royal Society open science},
  volume={4},
  number={11},
  pages={170623},
  year={2017},
  publisher={The Royal Society Publishing}
}

@article{henrique2019literature,
  title={Literature review: Machine learning techniques applied to financial market prediction},
  author={Henrique, Bruno Miranda and Sobreiro, Vinicius Amorim and Kimura, Herbert},
  journal={Expert Systems with Applications},
  volume={124},
  pages={226--251},
  year={2019},
  publisher={Elsevier}
}

@book{hamilton2020time,
  title={Time series analysis},
  author={Hamilton, James D},
  year={2020},
  publisher={Princeton university press}
}

@article{athey2019machine,
  title={Machine learning methods that economists should know about},
  author={Athey, Susan and Imbens, Guido W},
  journal={Annual Review of Economics},
  volume={11},
  number={1},
  pages={685--725},
  year={2019},
  publisher={Annual Reviews}
}

@article{siami2018forecasting,
  title={Forecasting economics and financial time series: ARIMA vs. LSTM},
  author={Siami-Namini, Sima and Namin, Akbar Siami},
  journal={arXiv preprint arXiv:1803.06386},
  year={2018}
}

@article{li2020multimodal,
  title={A multimodal event-driven LSTM model for stock prediction using online news},
  author={Li, Qing and Tan, Jinghua and Wang, Jun and Chen, Hsinchun},
  journal={IEEE Transactions on Knowledge and Data Engineering},
  volume={33},
  number={10},
  pages={3323--3337},
  year={2020},
  publisher={IEEE}
}

@article{10.1145/3688399,
author = {Xing, Frank},
title = {Designing Heterogeneous LLM Agents for Financial Sentiment Analysis},
year = {2024},
publisher = {Association for Computing Machinery},
address = {New York, NY, USA},
issn = {2158-656X},
url = {https://doi.org/10.1145/3688399},
doi = {10.1145/3688399},
note = {Just Accepted},
journal = {ACM Trans. Manage. Inf. Syst.},
month = aug
}

@inproceedings{10.1145/3637528.3671552,
author = {Liu, Zhiwei and Yang, Kailai and Xie, Qianqian and Zhang, Tianlin and Ananiadou, Sophia},
title = {EmoLLMs: A Series of Emotional Large Language Models and Annotation Tools for Comprehensive Affective Analysis},
year = {2024},
isbn = {9798400704901},
publisher = {Association for Computing Machinery},
address = {New York, NY, USA},
url = {https://doi.org/10.1145/3637528.3671552},
doi = {10.1145/3637528.3671552},
booktitle = {Proceedings of the 30th ACM SIGKDD Conference on Knowledge Discovery and Data Mining},
pages = {5487–5496},
numpages = {10},
location = {Barcelona, Spain},
series = {KDD '24}
}

@article{huang2024scalability,
  title={Scalability and Security of Blockchain-empowered Metaverse: A Survey},
  author={Huang, Huawei and Yin, Zhaokang and Yang, Qinglin and Li, Taotao and Luo, Xiaofei and Zhou, Lu and Zheng, Zibin},
  journal={IEEE Open Journal of the Computer Society},
  year={2024},
  publisher={IEEE}
}

@article{truong2023blockchain,
  title={Blockchain meets metaverse and digital asset management: A comprehensive survey},
  author={Truong, Vu Tuan and Le, Long and Niyato, Dusit},
  journal={Ieee Access},
  volume={11},
  pages={26258--26288},
  year={2023},
  publisher={IEEE}
}

@article{le2024blockchain,
  title={Blockchain-Empowered Metaverse: Decentralized Crowdsourcing and Marketplace for Trading Machine Learning Data and Models},
  author={Le, Hung Duy and Truong, Vu Tuan and Le, Long Bao},
  journal={IEEE Access},
  year={2024},
  publisher={IEEE}
}

@inproceedings{lee2024test,
  title={A Test Method for the Convergence of the Metaverse and Blockchain},
  author={Lee, Tae-Gyu},
  booktitle={2024 26th International Conference on Advanced Communications Technology (ICACT)},
  pages={321--326},
  year={2024},
  organization={IEEE}
}

\appendices

\section{Detailed Contributions to Related Work}
\label{app:contributions}

This study advances four interconnected research streams.

\textbf{Metaverse and Decentraland Research.} 
While prior work has established the technological foundations of the Metaverse \citep{duan2021metaverse} and Decentraland's specific implementation \citep{guidi2022social, den2023metaverse}, this study is among the first to empirically link platform-specific community sentiment to token price dynamics (see Table~\ref{tab:comparison} for performance metrics). We demonstrate that DAO-governed virtual worlds \citep{wu2022metaverse} generate user-generated content \citep{cheong2008consumers, bahtar2016impact} with measurable but bounded predictive signal for MANA returns, extending research on blockchain-empowered metaverse economies \citep{le2024blockchain} and the importance of monitoring user sentiment for platform governance \citep{micu2017analyzing, yang2015twitter}. The daily sentiment trajectory shown in Figure~\ref{fig:daily_sentiment} illustrates the temporal dynamics of community engagement.

\textbf{Sentiment Analysis Methodology.} 
Our work bridges lexicon-based and machine-learning approaches \citep{devika2016sentiment, gonccalves2013comparing, medhat2014sentiment} with modern LLM applications. By validating that general-domain sentiment surveys \citep{wankhade2022survey, hussein2018survey, yue2019survey, neri2012sentiment} require domain adaptation for metaverse contexts, we contribute to methodological refinements in social media sentiment analysis \citep{zimbra2018state, melton2021public, alharbi2019twitter} and crypto-specific opinion mining \citep{chen2019sentiment, quan2023decoding}. The textual data structure summarized in Table~\ref{tab:textual_data} enables these analyses.

\textbf{LLM Applications in Financial Forecasting.} 
We advance the deployment of large language models \citep{kasneci2023chatgpt, chang2024survey, huang2023chatgpt, li2018deep} for specialized financial sentiment tasks. Utilizing architectures like BERT and RoBERTa \citep{liu2019roberta, xu2019bert, zhan2024optimization}, we test their efficacy on Discord data, contributing to research on LLM-based social media simulation \citep{tornberg2023simulating} and Reddit-style market sentiment analysis \citep{deng2023llms} within decentralized virtual communities.

\textbf{Cryptocurrency Prediction and Volatility.} 
Addressing the evolutionary dynamics of crypto markets \citep{elbahrawy2017evolutionary}, our LSTM-based multi-modal framework responds to calls for advanced prediction methodologies \citep{henrique2019literature}. We contribute to time series forecasting literature \citep{hamilton2020time, siami2018forecasting} by quantifying the marginal value of sentiment features beyond technical indicators \citep{athey2019machine}, demonstrating both the potential and limitations of community-specific data for crypto-volatility modeling. 

\section{Data Specifications and Empirical Results}
\label{app:data}

We detail the dataset structures and empirical benchmarks supporting the multi-modal analysis.

\subsection{Market Data Structure}
Table~\ref{tab:numerical_data} defines the quantitative financial variables extracted from Decentraland's MANA token markets. These time-series features constitute the baseline price-only model inputs.
\begin{table}[htbp]
  \caption{Numerical Data: MANA Price, Volume, Market Cap}
  \label{tab:numerical_data}
  \centering
  \renewcommand{\arraystretch}{1.3}
  \setlength{\tabcolsep}{12pt}
  \begin{tabular}{@{}llc@{}}
    \toprule
    \rowcolor{headerblue}
    \textcolor{white}{\textbf{Variable}} & 
    \textcolor{white}{\textbf{Description}} & 
    \textcolor{white}{\textbf{Unit}} \\
    \midrule
    Date & Trading timestamp & --- \\
    \rowcolor{zebra}
    Open Price & Opening market price & \faDollarSign\ USD \\
    High Price & Daily maximum price & \faDollarSign\ USD \\
    \rowcolor{zebra}
    Low Price & Daily minimum price & \faDollarSign\ USD \\
    Close Price & Closing market price & \faDollarSign\ USD \\
    \rowcolor{zebra}
    Volume & Trading volume (tokens exchanged) & \faCoins\ MANA \\
    Market Cap & Total market capitalization & \faDollarSign\ USD \\
    \bottomrule
  \end{tabular}
  \\[6pt]
  \footnotesize
  \textbf{Note:} All price and market cap values are denominated in US Dollars.
\end{table}

\subsection{Sentiment Corpus Structure}
Table~\ref{tab:textual_data} outlines the Discord message schema. The sentiment labels and scores are derived via LLM-based classification of the Content field.
\begin{table}[htbp]
  \caption{Textual Data: Users' Messages (Discord Corpus)}
  \label{tab:textual_data}
  \centering
  \renewcommand{\arraystretch}{1.3}
  \setlength{\tabcolsep}{10pt}
  \begin{tabular}{@{}llc@{}}
    \toprule
    \rowcolor{headerblue}
    \textcolor{white}{\textbf{Field}} & 
    \textcolor{white}{\textbf{Description}} & 
    \textcolor{white}{\textbf{Type}} \\
    \midrule
    Date \& Time & Message timestamp (UTC) & \faCalendar\ \faClock \\
    \rowcolor{zebra}
    Author ID & Anonymized user identifier & \faUserSecret \\
    Content & Raw message text body & \faComment \\
    \rowcolor{zebra}
    Attachments & Media/files shared & \faPaperclip \\
    Reactions & Emoji responses count & \faSmile[regular] \\
    \rowcolor{zebra}
    Label & Assigned sentiment category & \faTag \\
    Score & Numerical sentiment polarity ($-1$ to $+1$) & \faChartLine \\
    \bottomrule
  \end{tabular}
  \\[6pt]
  \footnotesize
  \textbf{Note:} Sentiment labels and scores derived via LLM-based classification.
\end{table}

\subsection{Model Performance Benchmarks}
Table~\ref{tab:comparison} presents the eight-fold cross-validation results comparing price-only LSTM baselines against multi-modal variants (Price + Sentiment). The average improvements indicate that sentiment features provide marginal but consistent predictive enhancement, though negative $R^2$ values in both configurations suggest substantial unmodeled volatility drivers.

\definecolor{headerblue}{RGB}{65,105,225}    
\definecolor{lightyellow}{RGB}{255,250,205}  
\definecolor{goodgreen}{RGB}{0,128,0}        
\definecolor{badred}{RGB}{220,20,60}         

\begin{table}[t]
  \caption{Comparison of Prediction Models (8 Independent Runs)}
  \label{tab:comparison}
  \centering
  \footnotesize
  \setlength{\tabcolsep}{4pt}
  \renewcommand{\arraystretch}{1.25}  
  \begin{tabular}{@{}c >{\raggedright\arraybackslash}p{1.5cm} ccc@{}}
    \toprule
    \rowcolor{headerblue}
    \textcolor{white}{\textbf{Run}} & 
    \textcolor{white}{\textbf{Model}} & 
    \textcolor{white}{\textbf{MSE ($\times 10^{-4}$)}} & 
    \textcolor{white}{\textbf{MAE}} & 
    \textcolor{white}{\textbf{R\textsuperscript{2}}} \\
    \midrule
    \multirow{2}{*}{1} 
      & Price Only & 5.50 & 0.01835 & $-$0.192 \\
      & +Sentiment & 5.53\textcolor{badred}{$\uparrow$} & 0.01816\textcolor{goodgreen}{$\downarrow$} & $-$0.198\textcolor{badred}{$\downarrow$} \\
    \multirow{2}{*}{2} 
      & Price Only & 5.30 & 0.01840 & $-$0.148 \\
      & +Sentiment & 5.27\textcolor{goodgreen}{$\downarrow$} & 0.01949\textcolor{badred}{$\uparrow$} & $-$0.141\textcolor{goodgreen}{$\uparrow$} \\
    \multirow{2}{*}{3} 
      & Price Only & 5.79 & 0.01853 & $-$0.256 \\
      & +Sentiment & 5.57\textcolor{goodgreen}{$\downarrow$} & 0.01784\textcolor{goodgreen}{$\downarrow$} & $-$0.207\textcolor{goodgreen}{$\uparrow$} \\
    \multirow{2}{*}{4} 
      & Price Only & 5.79 & 0.01853 & $-$0.256 \\
      & +Sentiment & 5.57\textcolor{goodgreen}{$\downarrow$} & 0.01784\textcolor{goodgreen}{$\downarrow$} & $-$0.207\textcolor{goodgreen}{$\uparrow$} \\
    \multirow{2}{*}{5} 
      & Price Only & 6.75 & 0.01981 & $-$0.463 \\
      & +Sentiment & 5.26\textcolor{goodgreen}{$\downarrow$} & 0.01916\textcolor{goodgreen}{$\downarrow$} & $-$0.140\textcolor{goodgreen}{$\uparrow$} \\
    \multirow{2}{*}{6} 
      & Price Only & 5.96 & 0.01870 & $-$0.291 \\
      & +Sentiment & 5.27\textcolor{goodgreen}{$\downarrow$} & 0.01807\textcolor{goodgreen}{$\downarrow$} & $-$0.143\textcolor{goodgreen}{$\uparrow$} \\
    \multirow{2}{*}{7} 
      & Price Only & 6.34 & 0.01913 & $-$0.375 \\
      & +Sentiment & 5.46\textcolor{goodgreen}{$\downarrow$} & 0.01932\textcolor{badred}{$\uparrow$} & $-$0.184\textcolor{goodgreen}{$\uparrow$} \\
    \multirow{2}{*}{8} 
      & Price Only & 5.31 & 0.01818 & $-$0.150 \\
      & +Sentiment & 5.24\textcolor{goodgreen}{$\downarrow$} & 0.01925\textcolor{badred}{$\uparrow$} & $-$0.136\textcolor{goodgreen}{$\uparrow$} \\
    \midrule
    \rowcolor{lightyellow}
    \multirow{-2}{*}{\textbf{Avg}}  
      & \textbf{Price Only} & \textbf{5.84} & \textbf{0.0187} & \textbf{$-$0.266} \\
    \rowcolor{lightyellow}
      & \textbf{+Sentiment} & \textbf{5.40}\textcolor{goodgreen}{\textbf{$\downarrow$}} & \textbf{0.0186}\textcolor{goodgreen}{\textbf{$\downarrow$}} & \textbf{$-$0.170}\textcolor{goodgreen}{\textbf{$\uparrow$}} \\
    \bottomrule
  \end{tabular}
  \\[4pt]
  \scriptsize
  \textbf{Note:} MSE scaled by $10^{4}$ for readability. Legend: \textcolor{goodgreen}{$\downarrow$}=Lower error/higher R\textsuperscript{2} (better); \textcolor{badred}{$\uparrow$}=Worse.
\end{table}

\subsection{Temporal Visualization}
\begin{figure}[htbp]
  \centering
  \fcolorbox{headerblue}{white}{\includegraphics[width=0.98\linewidth]{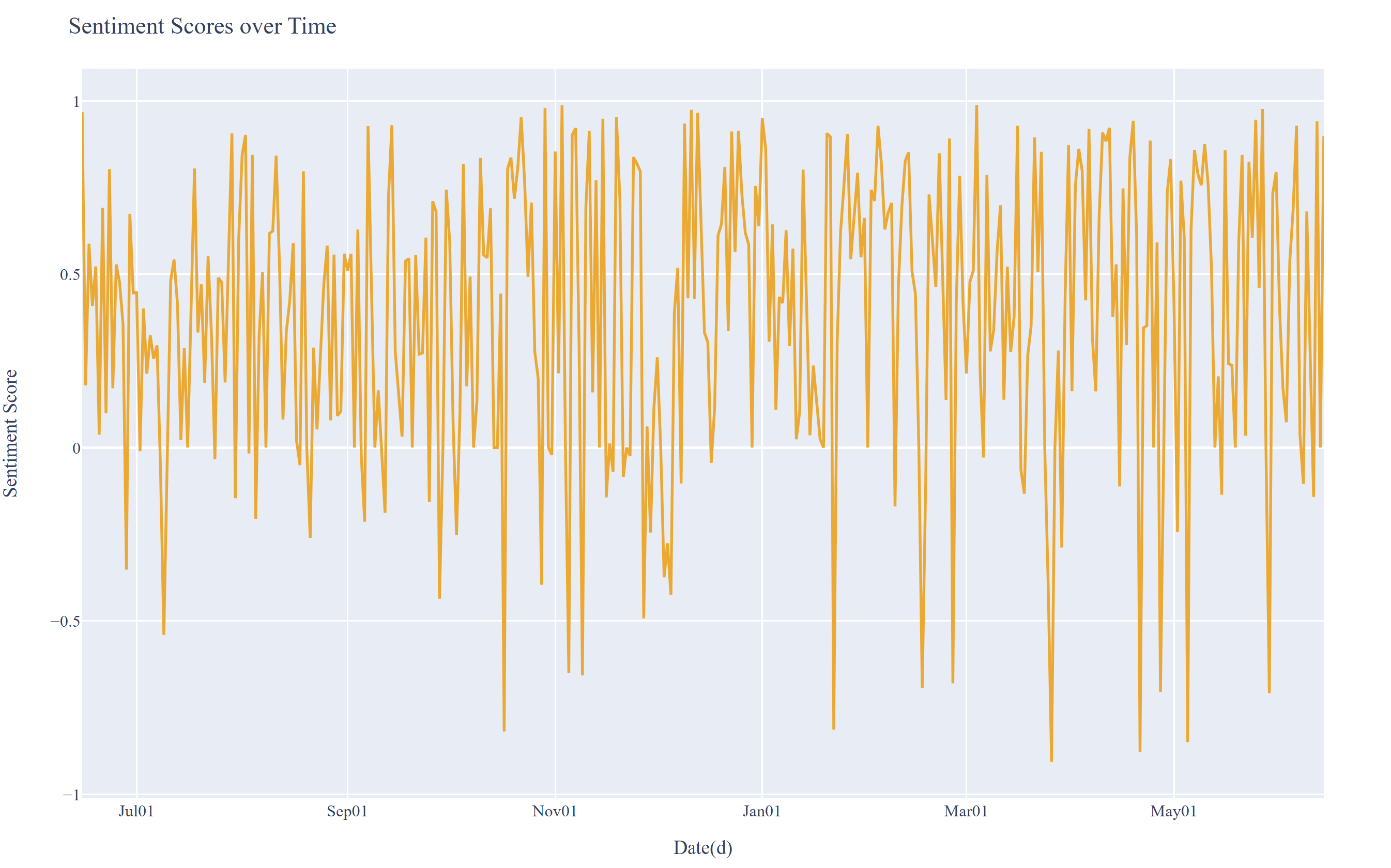}}
  \captionsetup{skip=0.1cm, font=footnotesize}
  \caption{Daily Average Sentiment Score (Decentraland Discord Community).}
  \label{fig:daily_sentiment}
\end{figure}

\begin{figure}[htbp]
\centering
 \fcolorbox{headerblue}{white}
{\includegraphics[width=1\linewidth]{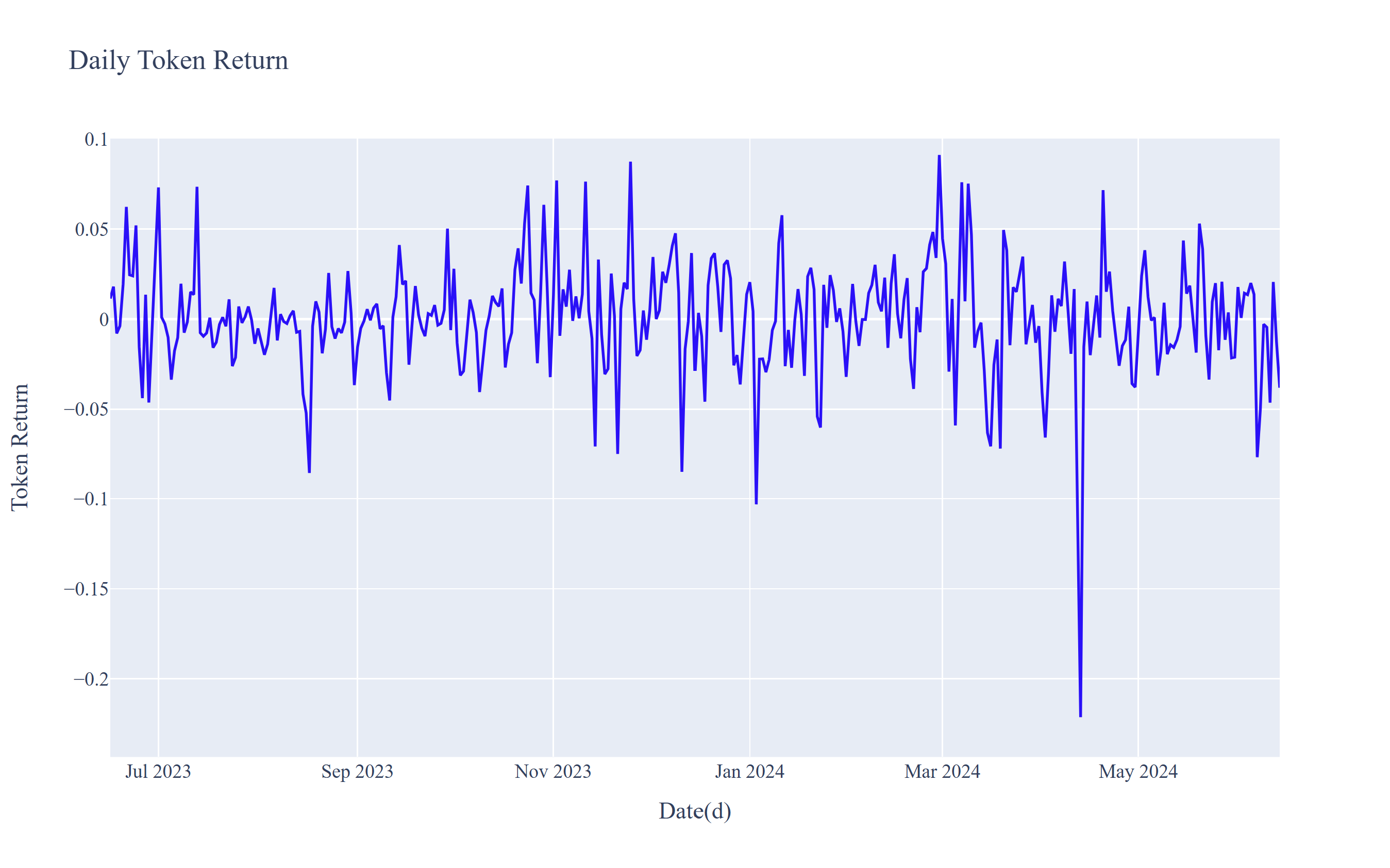}}
\captionsetup{skip=0.1cm}
\captionsetup{font=footnotesize}
\caption{Daily Token Return.}
\label{fig: daily-token-return}
\end{figure}

Figure~\ref{fig:daily_sentiment} displays the daily averaged sentiment scores derived from the Discord corpus (Table~\ref{tab:textual_data}), revealing temporal fluctuations in community mood. Figure~\ref{fig: daily-token-return} overlays predicted returns from the multi-modal model against realized returns, visualizing the systematic underestimation of volatility discussed.





\end{document}